\documentclass[letterpaper, 10 pt, conference]{ieeeconf}

\IEEEoverridecommandlockouts  
\usepackage[colorlinks=true, allcolors=blue]{hyperref}
\usepackage{graphics}
\usepackage{amsmath}
\usepackage{amssymb}
\usepackage{comment}
\usepackage{todonotes}
\presetkeys{todonotes}{inline,backgroundcolor=pink,caption={}}{}

\usepackage{siunitx}
\usepackage{balance}
\usepackage{soul}
\usepackage{csquotes}
\usepackage[english]{babel}
\addto\captionsenglish{\renewcommand{\figurename}{Fig.}} %

\title{ \bf \LARGE 
A High-Force Gripper with Embedded Multimodal Sensing for Powerful and Perception Driven Grasping
}

\author{Edoardo\,Del\,Bianco$^{\dagger\S}$,
Davide\,Torielli$^{\ddagger}$,
Federico\,Rollo$^{\dagger\S}$,
Damiano\,Gasperini$^{\ddagger\#}$,
Arturo\,Laurenzi$^{\ddagger}$,\\
Lorenzo\,Baccelliere$^{\ddagger}$,
Luca\,Muratore$^{\ddagger}$,
Marco\,Roveri$^{\S}$,
and Nikos\,G.\,Tsagarakis$^{\ddagger}$
\thanks{This work was supported by the Leonardo S.p.A under the grant LDO/CTI/P/0020481/23 with Università di Trento, by the EU Horizon Europe project HARIA under the grant No.101070292 with Istituto Italiano di Tecnologia and by the EU Horizon Europe CONCERT under the grant No.101016007 with Istituto Italiano di Tecnologia}
\thanks{$^{\dagger}$Leonardo Innovation Labs, Leonardo S.p.A., Genoa, Italy}
\thanks{\ \ \footnotesize e-mail: {\tt\footnotesize\{name.surname\}@leonardo.com}}
\thanks{$^{\ddagger}$HHCM, Istituto Italiano di Tecnologia, Genoa, Italy}
\thanks{\ \ \footnotesize e-mail: {\tt\footnotesize\{name.surname\}@iit.it}}
\thanks{$^{\S}$Industrial Innovation, DISI, Università di Trento, Trento, Italy}
\thanks{\ \ \footnotesize e-mail: {\tt\footnotesize\{name.surname\}@unitn.it}}
\thanks{$^{\#}$DIBRIS, Università degli Studi di Genova, Genova, Italy}
}

\begin{document}

\maketitle
\thispagestyle{empty}
\pagestyle{empty}

\begin{abstract}
Modern humanoid robots have shown their promising potential for executing various tasks involving the grasping and manipulation of objects using their end-effectors. Nevertheless, in the most of the cases, the grasping and manipulation actions involve low to moderate payload and interaction forces. This is due to limitations often presented by the end-effectors, which can not match their arm-reachable payload, and hence limit the payload that can be grasped and manipulated. In addition, grippers usually do not embed adequate perception in their hardware, and grasping actions are mainly driven by perception sensors installed in the rest of the robot body, frequently affected by occlusions due to the arm motions during the execution of the grasping and manipulation tasks.
To address the above, we developed a modular high grasping force gripper equipped with embedded multi-modal perception functionalities. The proposed gripper can generate a grasping force of 110 N in a compact implementation. The high grasping force capability is combined with embedded multi-modal sensing, which includes an eye-in-hand camera, a Time-of-Flight (ToF) distance sensor, an Inertial Measurement Unit (IMU) and an omnidirectional microphone, permitting the implementation of perception-driven grasping functionalities. 

We extensively evaluated the grasping force capacity of the gripper by introducing novel payload evaluation metrics that are a function of the robot arm's dynamic motion and gripper thermal states. We also evaluated the embedded multi-modal sensing by performing perception-guided enhanced grasping operations. 

\end{abstract}

\section{Introduction} \label{sec:introduction}
As technology progresses and humanoid robots become more mature, grasping systems play a critical role in enabling the development of their manipulation capabilities. Traditionally, humanoids have been equipped with robotic hands inspired by the human hand. These systems have been deployed with success even in extreme environments like the deep sea \cite{brantner2021controlling}. Nevertheless, the efficacy of innovative approaches, like the jamming of granular material \cite{brown2010universal}, drives us to reflect on the underlying principles behind the grasping operation.
Langowski et al.\ looked at nature as a source of inspiration for new grasping paradigms \cite{langowski2020soft}. With this work, the authors observed the methods which animals employ to grasp objects around them, and which level of power, dexterity and compliance they can achieve. These observations can be extended to dogs, which picks up objects with their mouths, instead of their paws. Experience suggests that dogs obtain a very stable and secure grasp of various objects just with their bite. The force they exert is trimmed singularly according to what the dog is carrying or pulling (e.g., a rolled newspaper, a heavy toy or a rug), excelling in high-force grasp at the cost of dexterity. Hence, a dog's mouth can be assimilated to a gripper with two jaws, one fixed and one revolving and powered by an actuator that can regulate the grasping force. This observation inspired the design of the gripper introduced in this paper.

\begin{figure}
    \centering
\includegraphics[width=0.95\linewidth,trim={4cm 2cm 2cm 0},clip]{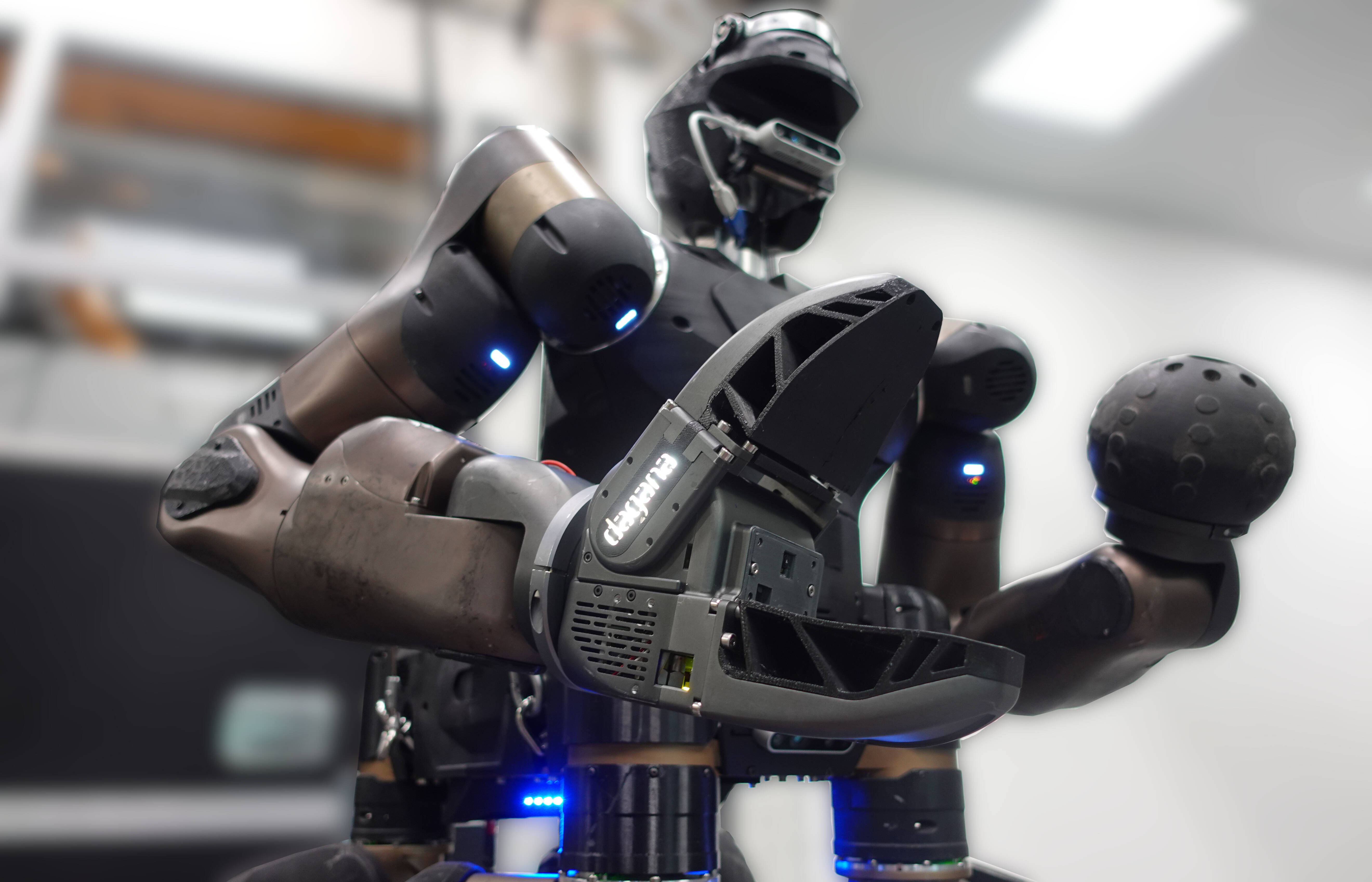}
    \caption{The developed gripper mounted on the arm of the CENTAURO robot.}
    \label{fig:photo_description_1}
\end{figure}

We developed the proposed gripper for performing high-force, perception-enhanced grasping operations with high-power humanoid robots, like the CENTAURO platform~\cite{kashiri2019centauro} (\figurename{}~\ref{fig:photo_description_1}). The gripper demonstrates the following innovative characteristics:
\begin{itemize}
    \item Its single revolute joint is powered by a high-power torque-controlled Series Elastic Actuator (SEA), allowing the impedance modulation and control of its grasp with torques up to \SI{16.5}{\newton\meter} that is translated to forces up to \SI{110}{\newton} at \SI{0.15}{\meter} distance from the gripper joint.
    \item Its body has been realized in a modular way, including the main body, jaws, pads, and sensing modules, which permit the quick adaptation of the gripper form, matching the requirements of a variety of grasping tasks.
    \item It features proprioceptive and exteroceptive sensing capabilities (video, audio, distance measurement and inertial measurement) that enable the implementation of precise and perception-driven autonomously controlled grasping operations. Such perception-driven actions implemented include vision-guided and artificial-sensing-enhanced grasping skills.
    \item Finally, in the assessment of the gripper grasping force capacity, rather than evaluating the standard static grasping payload performance, novel grasping payload evaluation metrics are introduced in which the robot arm dynamic motion and gripper thermal states are taken into account to reveal the functional range of gripper grasping force under dynamic arm motion and gripper actuation thermal stressing conditions. 
    
\end{itemize}

We demonstrated the features of the developed gripper in a series of mechatronic tests and perception-driven grasping tasks when equipped on the CENTAURO robot~\cite{kashiri2019centauro}, a high-payload quadruped robot with a humanoid upper body.

The structure of the article is divided as follows. In Section \ref{sec:related_works}, the related works are presented, highlighting the distinctive features of the developed gripper with respect to the state-of-the-art. The gripper design is presented in Section \ref{sec:device_description}. In Section \ref{sec:percGrasp}, the perception-driven grasping abilities of the gripper are described. In Section \ref{sec:experiments}, experiments are carried out to validate its mechatronic design, grasping force capacity, and perception-enhanced functionalities. Finally, in Section \ref{sec:conclusions}, conclusions are drawn.

\section{Related works} \label{sec:related_works}
The design of grasping systems can vary with different grasping transmissions (such as direct, cable, compressed air), grasping bodies (like claws, fingers, suction cups), and actuation paradigms (like electric, pneumatic and hydraulic)~\cite{gripperDesignSurvey, samadikhoshkho2019brief}, with each category showing different capabilities and adaptability to various kinds of tasks and environments~\cite{Negrello2020}.

Simple 2-pads, often parallel, grippers are one of the most common choices for robots, with the advantages of simplicity, cost-effectiveness, and usually good power-to-weight ratio. This preference is also demonstrated by the high number of commercial solutions available~\cite{torielli2023ros}. Such designs, with only one actuation degree, decrease the chance of failure and increase the overall gripper robustness~\cite{ma2016m}.

Despite the spread of 2-pads grippers, everyday objects are modeled according to the much more articulated human hand, requiring a more dexterous robotic end-effector to deal with them, designed with a fingered grasping mechanism~\cite{Piazza2019}. Such systems can range from including a few under-actuated fingers~\cite{Catalano2016, RBOHand22016, ren2018heri, Liu2020}, to present fully-actuated, anthropomorphic hand~\cite{AllegroHand2012, Ruehl2014, shadowhand2014}.
Despite under-actuated finger solutions increase the adaptability to the object, they add complexity, resulting in less robust mechanisms that cannot reach the strength of 2-pads grippers. Similarly, fully articulated human-like hands can show a very high dexterity for manipulation tasks, but their higher complexity adds more costs, with less robustness and strength. Furthermore, the presence of a complex actuator and transmission system also leads to bulkier and heavier bodies, limiting their integration into real-world applications.
The developed gripper belongs to the 2-body gripper category but with a series of novel features. Its actuation system shows a higher strength-to-weight ratio than the existing solutions of this category~\cite{wu2023back}.

Another interesting end-effector design choice is the modularity of the system.
To provide a flexible system that can be adapted to a greater variety of tasks, modularity is employed in the finger-hand assembly, extending the capabilities of these grippers and easing the fabrication and assembly of the parts. For example, fingers can be added, removed or rearranged~\cite{ren2018heri}, and phalanges modified~\cite{Chao2023}.
Changing the position and shape of one of the fingers increases the capabilities of grippers allowing them to grasp objects of various size~\cite{maeda2022f1, telegenov2015low}.
In the design choices of the proposed gripper, we included different levels of modularity. Indeed, the two jaw modules can be quickly replaced to address new use cases requiring different jaw shapes and materials optimized to accommodate grasping a particular tool or object. Modularity is also considered for the embedded perception since the sensing module can be easily removed and replaced with others.
Thus, even if the grasping jaw grasping system design of the proposed is already present in other solutions, e.g., on the end-effector mounted on the Boston Dynamics Spot quadruped robot~\cite{bostondynamics2024}, compared to these existing grippers, the proposed end-effector offers not only extended grasping force capacity but also higher reconfigurability thanks to its modular grasping and perception modules.

Together with kinematic and actuation design, the research has also focused on equipping perception capabilities in the gripper itself. Exteroceptive (like tactile sensors, accelerometers, and cameras) and proprioceptive sensors (like motor encoders, and certain types of force/torque sensors) offer remarkable possibilities in robotic grasping. For example, tactile sensing~\cite{KAPPASSOV2015195} helps in applying a specific grasping force~\cite{Romano2011, yamaguchi2016combining}, or detect slipping~\cite{Narita2020}, while eye-in-hand vision systems can focus on the object while the arm is moving, demonstrating impressive use cases, like catching objects~\cite{cigliano2015robotic, Hundhausen2021}.
Also, embedded range sensors have shown their potentiality in the grasping task~\cite{Cirillo2021}.
Concerning microphones, recent works show how visual and audio data can be fused to improve the manipulation of objects~\cite{Mejia2024}.
In general, it is demonstrated the importance not only to equip such sensors but also to integrate them in the end-effector capabilities, to allow for really empowering such robotic systems~\cite{torielli2023ros}. %
The gripper proposed in this work integrates embedded multi-modal perception that combines visual and distal sensing to explore the multi-modal perception cues for the development of autonomous grasping functionalities. Furthermore, thanks to audio sensing capabilities, it provides a seamless human-robot vocal interaction thanks to the recognition of the user's vocal commands. To the best of the authors' knowledge, no such system integrates all these sensing capabilities directly on the end-effector itself and provides a human-interaction perception grasping pipeline which combines all the potentialities of these sensors.

Once an end-effector is developed, it is important to understand its capabilities. 
Indeed, some studies introduce benchmarks for evaluating the end-effector dexterity~\cite{Denoun2023}, regarding manipulating everyday objects~\cite{Sotiropoulos2018} and  clothes~\cite{Angus2023}. Other works focus more on intrinsic benchmarks, providing standards for measuring parameters like finger-applied forces and grasping strength~\cite{Falco2015}.
Nevertheless, works that introduce new end-effector systems often lack such deeply evaluations~\cite{AllegroHand2012, Ruehl2014, shadowhand2014, Catalano2016, RBOHand22016, ren2018heri, Liu2020, maeda2022f1, Chao2023} and they do not delve into evaluating properly the strength and power capacity of the developed gripper, describing, when present, only the maximum static payload. This limited amount of information also affects the data sheets of commercial solutions (e.g., Schunk, OnRobot and Robotiq). %
In our work,  we evaluate more extensively the grasping force capacity of the developed gripper (Section~\ref{sec:experiments_mech}), delving into more deep grasping force tests, showing not only static grasped payload capacity, but also other metrics like the relation between applied torque and tips force, how the actuator's temperature rises, and the dynamic payload capacity (i.e., considering the acceleration of the grasped payload due to the arm motion). Despite the range of possible tests can grow up indefinitely, the presented ones allow to fully understand the capabilities and limits of the presented gripper, differently from single value specification of the maximum grasping force and payload that is common to see in previous works and data sheet of most gripper devices.

\section{Device Description} \label{sec:device_description}
The gripper device presented in this work combines two main features: \textit{(i)} high grasping force capacity and \textit{(ii)} multi-modal sensing. These functionalities enable two capabilities: \textit{(i)} the execution of grasping tasks demanding human-scale grasping forces while, at the same time, \textit{(ii)} permitting to realize autonomous grasping principles thanks to the embedded sensing subsystems. The aforementioned capabilities are considered fundamental for executing physically robust real-world grasping tasks that can be also carried out with a certain level of autonomy. 

\subsection{Mechatronic design}
\begin{figure}
    \centering
    \includegraphics[width=1\linewidth]{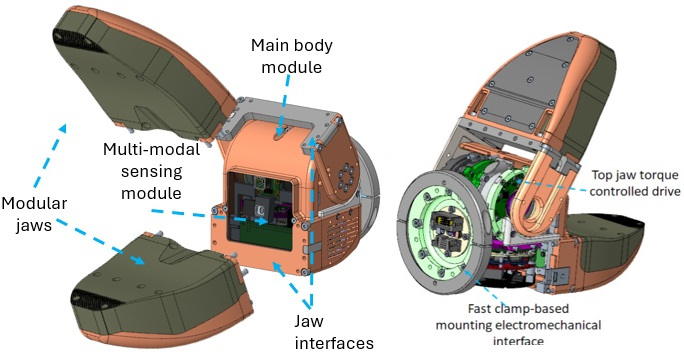}
    \caption{The CAD design of the gripper, highlighting its core modules.}
    \label{fig:cad_modules}
\end{figure}

The proposed gripper is fully modular, enabling independent development of its body elements and simplified task-driven physical reconfiguration of its hardware. 
Its three functional modules are shown in \figurename{}~\ref{fig:cad_modules} and can be quickly connected with a few fasteners. We can identify \textit{(i)} the main body module of the gripper, \textit{(ii)} the modular jaws, which connect to the main body module, and \textit{(iii)} the sensing module that is housed as an additional module on the front side of the main body module.
The combined weight of the main body and sensing modules is \SI{1.5}{\kilo\gram}, while the jaw modules weight \SI{0.35}{kg} each, giving a total weight, in the considered configuration, of \SI{2.2}{kg}.

\subsubsection{Main body module}
\begin{figure}
    \centering
    \includegraphics[width=1\linewidth]{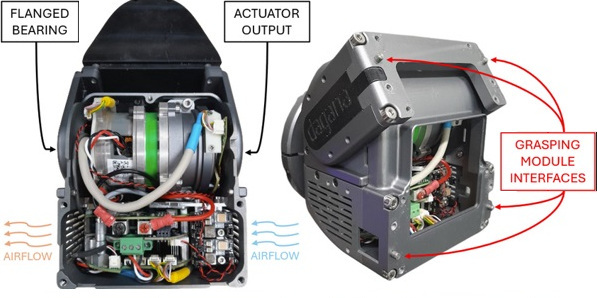}
    \caption{The gripper's actuator with its driver electronics.}
    \label{fig:photo_actuation}
\end{figure}

The high-force grasping gripper is powered by a series elastic torque-controlled actuator connected to the top moving jaw interface and housed inside the main body module, as shown in \figurename{}~\ref{fig:photo_actuation}. The actuator provides torque sensing, a continuous torque of approximately \SI{10}{\newton\meter}, and a peak torque of \SI{16.5}{\newton\meter}, leading to a peak pinching force of \SI{110}{\newton} at \SI{0.15}{\meter} distance from the center of the joint of the jaw. 

The main body module features a closed cell structure which encloses the actuator, its power and communication electronics and the sensing module, supporting and protecting them from impacts while maintaining a low weight. A forced wind tunnel from the sides of the main body permits airflow through the actuator body and its electronics. 

The torque-sensing capability of the actuator allows precise force control during the grasping operation, as well as it permits to actively regulate the impedance parameters (stiffness and damping) of the moving jaw module. A conical clamp-based electromechanical interface (right image of \figurename{}~\ref{fig:cad_modules}) has been designed to enable fast mounting of the gripper main body module to the robot wrist.

\subsubsection{Grasping jaw module}
While 1-DoF revolute-joint grippers allow for a very robust and reliable operation, to which the proposed gripper device adds its high-force grasping, this actuation mechanism can be disadvantageous while grasping objects of irregular shape that require adapting or enveloping. 

\begin{figure}
    \centering
    \includegraphics[width=1\linewidth]{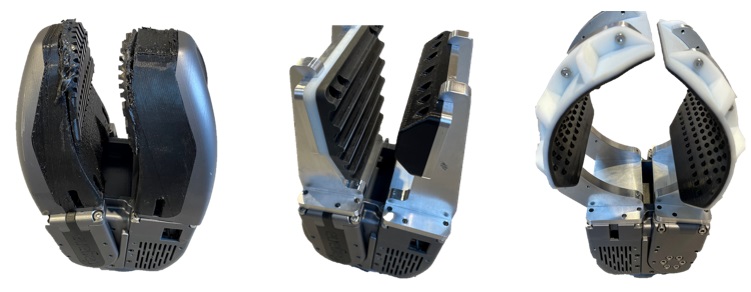}
    \caption{Different types of jaw modules, on the left general jaw modules, in the middle jaw modules dedicated to the grasping of planar formwork elements, and, on the right, jaw modules shaped to provide enhanced grasping of large cylindrical objects.}
    \label{fig:fem_grasping_module_examples}
\end{figure}

To partially overcome this weakness, common to this gripper type, we developed the jaw elements in a modular way, rendering them completely independent from the main body module of the gripper. Custom jaws can be quickly fitted as needed to facilitate specific grasping needs requested due to particular object shapes and sizes. Few examples of such grasping task-customized jaw modules are shown in \figurename{}~\ref{fig:fem_grasping_module_examples}.

The jaw modules are made from Aluminum\,7075-T6.
On the inner side of the jaws, 3D-printed rubber pads can be mounted. These pads can be easily replaced, modifying their shape and stiffness to obtain an effect of adaptability to the shape of the object.

\subsubsection{Gripper multi-modal sensing module}
Perception-enhanced grasping operations often rely on other sensors positioned in the main body of the robot, or externally in the environmental setup. This approach is not always suitable for perception-enhanced grasping tasks, due, for example, to varying occlusions due to the robot's motions. Therefore, we decided to incorporate sensing capabilities in the gripper itself, designing the main body module of the gripper with a housing in its front side for a multi-modal sensing unit.

The multi-modal sensing unit employed leverages on an Arduino Nicla Vision board, %
a single-board sensing and computation unit which features a \SI{2}{MP} RGB camera, a ToF (Time-of-Flight) distance sensor, an omnidirectional MEMS (Micro-Electro-Mechanical System) microphone, and a 6-axis IMU (Inertial Measurement Unit). The board also provides wireless communication to transmit the sensory data, eliminating the need of a wired-based communication interface. We selected this integrated sensing and perception board for multiple reasons:
\begin{itemize}
    \item Its overall dimensions permit its incorporation in the limited space available on the front side of the main body module of the gripper.
    \item The capability to communicate through Wi-Fi allows the use of the perception features while mounting the gripper on robots that are not provided with a wired data transmission line to the end-effector.
    \item Its onboard battery management system allows the installation of a small battery directly connected to the board, permitting the gripper to work on robots that do not have a dedicated power transmission line for the end-effector's sensing hardware. %
\end{itemize}

\begin{figure}
    \centering
    \includegraphics[width=1\linewidth]{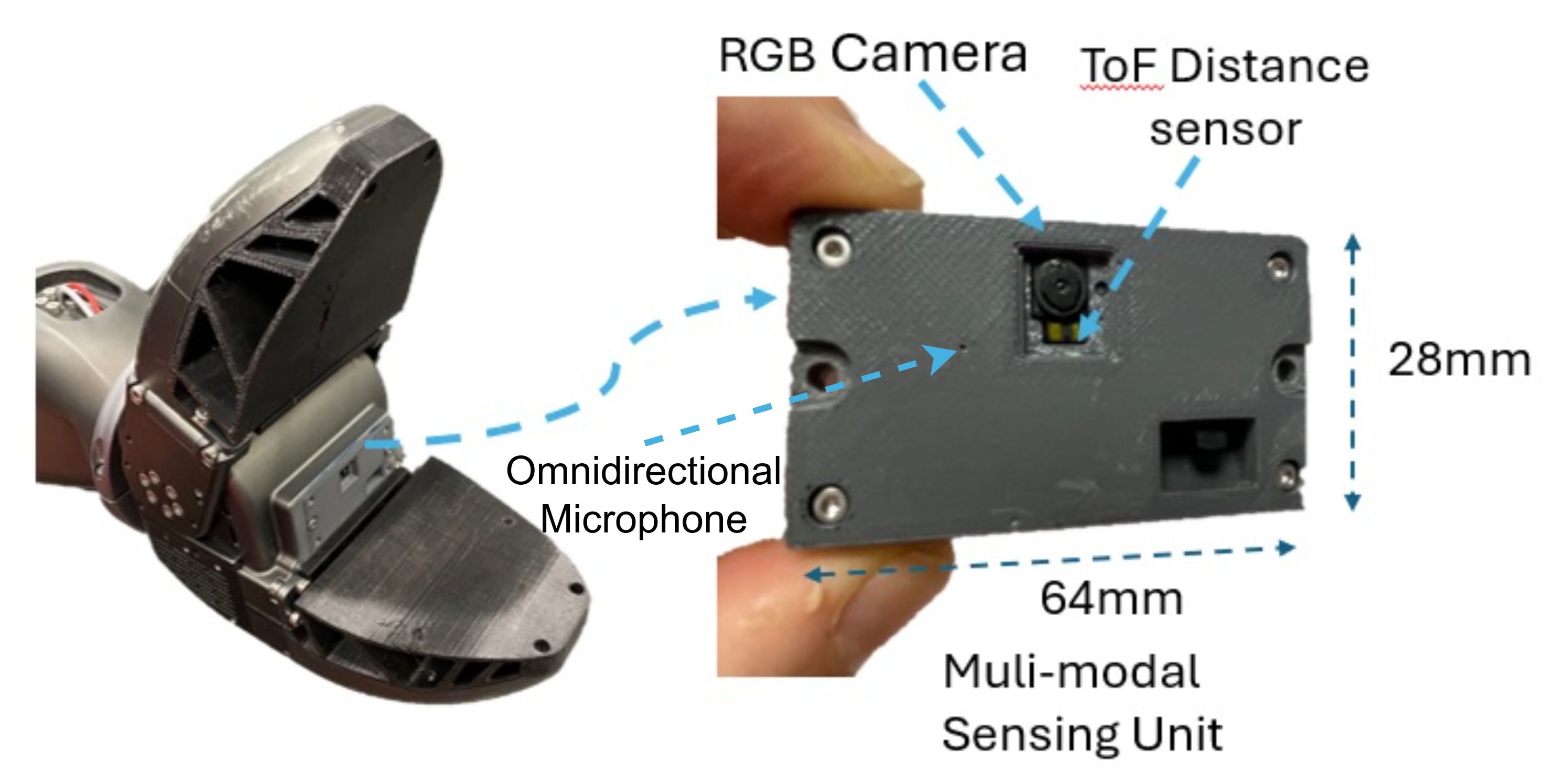}
    \caption{The add-on multi-modal sensing unit indicating indicating the principal sensors. The unit incorporates also a 6-axis IMU sensor.}
    \label{fig:cad_sensing_modules}
\end{figure}

We integrated the Nicla Vision perception unit with a \SI{3.7}{\volt} battery of \SI{820}{mAh}, a Wi-Fi antenna, and a power switch. Such assembly is placed into a perception unit enclosure that can be easily inserted and extracted from the front side of the main body module of the gripper, without the need to disassemble any other component (\figurename{}~\ref{fig:cad_sensing_modules}). A side benefit from this independent add-on perception unit setup is the opportunity to develop perception software without the need to use the whole gripper device until it is necessary. Thanks to the modular design, as new sensors become available, this unit can be upgraded without interfering with the other components of the gripper.

\subsection{Software Interfaces}

\subsubsection{Control Interface}
The low-level communication with the gripper electronics is made through EtherCAT\footnote{EtherCAT: \href{https://www.ethercat.org/it/technology.html}{https://www.ethercat.org/it/technology.html}} connection.
To fully exploit the grasping and sensing capabilities of the gripper, we implemented software modules in XBot~\cite{XBot2}, a real-time software architecture capable of interfacing with different kinds of robotic systems. XBot allows communicating seamlessly with the controlled robot, including the equipped end-effector(s), through ROS~\cite{ROS}, the de facto standard middleware for robotic applications. The implemented software tools also include functionalities to visualize and simulate the gripper in the standard ROS tools, RViz and Gazebo.

\subsubsection{Sensing Interface}
The software running on the multi-modal sensing module has been developed to provide a collection of the data from the camera, the distance sensor, the IMU and the microphone over Wi-Fi, with the possibility of choosing between UDP or TCP protocols from a configuration file. Wireless communication allows for fast prototyping and removes the necessity of additional cables. The communication with all the sensors is fully integrated with ROS and ROS2, allowing for seamless collection of the multi-modal sensing data. 
Furthermore, to complement the real robotic system, simulation tools have been developed to use the multi-modal sensing unit in Gazebo simulation.

\section{Perception Driven Grasping}\label{sec:percGrasp}

Thanks to its exteroceptive and proprioceptive sensing capabilities, autonomous grasping functionalities can be realized allowing the proposed gripper to detect, reach, and grasp objects of interest when commanded by the user. 

\begin{figure}
    \centering
\includegraphics[width=1\linewidth]{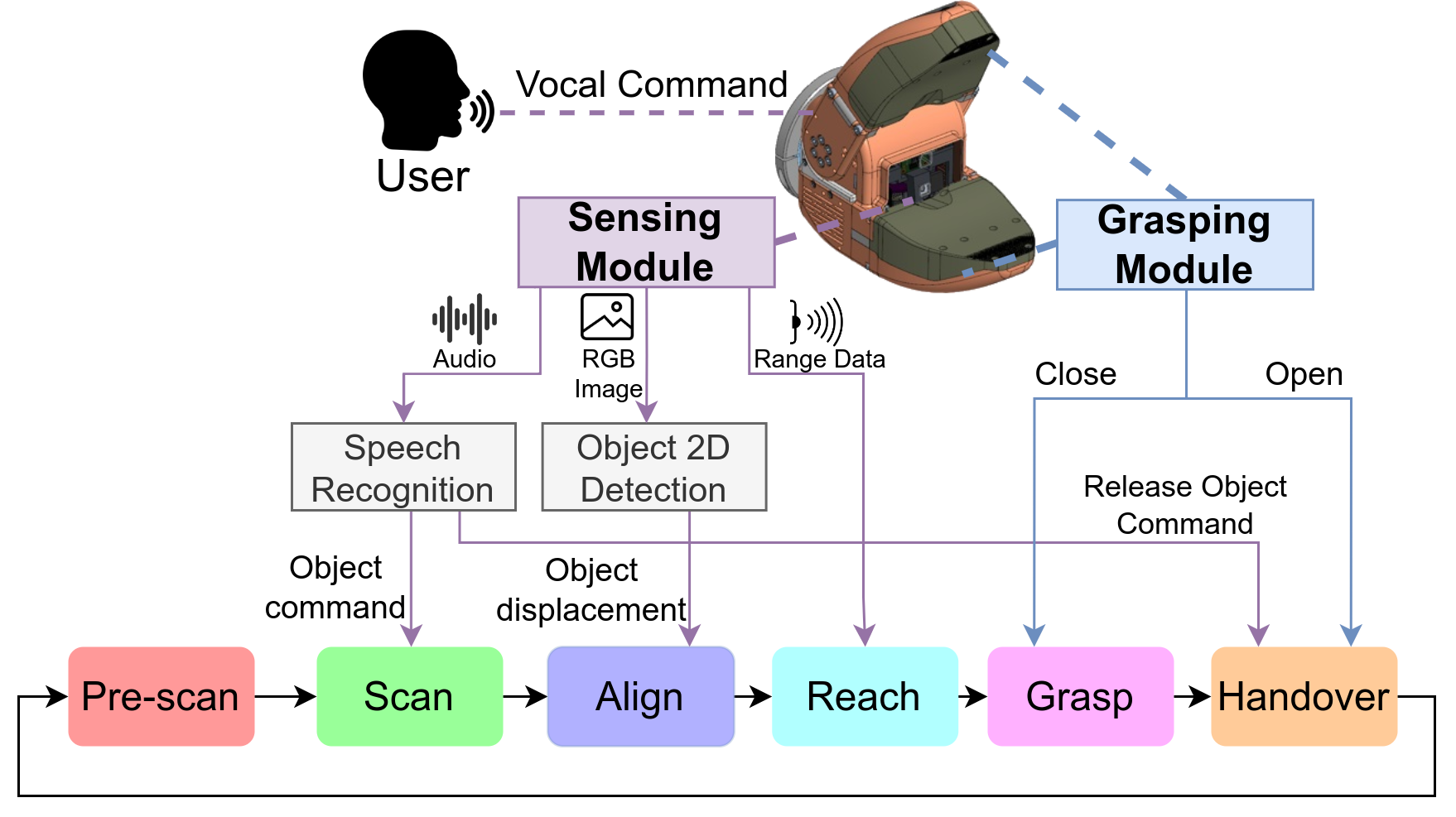}
    \caption{The gripper perception driven grasping pipeline. Combining the gripper sensing and actuation capabilities, a requested object can be reached and grasped.}
    \label{fig:graspPerceptionPipeline}
\end{figure}

For this purpose, a perception-driven grasping pipeline has been implemented, \figurename{}~\ref{fig:graspPerceptionPipeline}. The pipeline begins with an initial phase where the gripper looks for an object to be grasped in the environment with its embedded camera (\textit{Scan} phase). To extend the field of view, in this phase, the robot moves the wrist in order to search the environment. Meanwhile, the embedded microphone in the gripper listens to for user's commands, referring to which object to grasp. The user's vocal commands are recorded by the microphone, transmitted to the control computational unit of the robot, and fed to VOSK \cite{vosk}, an offline speech recognition toolkit. This toolkit recognizes the requested object category among the COCO classes~\cite{lin2014microsoft}. From this point on, the robot specifically searches the requested object. 

The detection of the object is achieved by feeding the images from the gripper's embedded camera to the YOLOv8 network\footnote{[Online]. Available: \href{https://github.com/ultralytics/ultralytics}{https://github.com/ultralytics/ultralytics}} (a network based on YOLO~\cite{redmon2016you}). Once the requested object is detected in the images, the \textit{Align} phase begins. The centroid of the object in the image $C$ is then computed by extracting the median point of the detected object bounding box. Following this, the horizontal angle $\theta_h$ and the vertical one $\theta_v$ are derived from the equation below:
\begin{equation}
    \theta_j = \frac{\mathit{fov}_j}{\mathit{dim}_j}(C_j - \mathit{ref}_j)
\end{equation}
where $j \in \{h, v\}$; $\mathit{fov}_j$ is the camera field of view; $\mathit{dim}_j$ are the dimensions of the image; $C_j$ are the current pixel values of the object center in the image; $\mathit{ref}_j$ is the pixel reference in which we want to move the object center, in our case it is $\mathit{ref}_j = \frac{\mathit{dim}_j}{2}$. 
The angles $\theta_h$ and $\theta_v$ are then used to align the gripper toward the object using a Cartesian PD controller as exposed in the following equation:
\begin{align}
    \begin{cases}
        x[i] = x[i-1]\\
        y[i] = y[i-1] + K_p \theta_h[i] + K_d (\theta_h[i] - \theta_h[i-1])\\
        z[i] = z[i-1] + K_p \theta_v[i] + K_d (\theta_v[i] - \theta_v[i-1])
    \end{cases}
\end{align}
considering that the reference frame, composed of the $x$, $y$, and $z$ axes, is the right-handed torso frame of the robot, which follows the standard convention of $x$ forward, $y$ left, and $z$ up. This regulation can be used with the assumption that the gripper's approach axis, i.e., the one that exits from the camera center, is almost parallel to the robot $x$-axis which is often true for these picking tasks.

Once the object is in the center of the camera, the \textit{Reach} phase is started, with the ToF sensor of the gripper multi-modal sensing module utilized to measure the distance between the gripper and the object, providing a rough estimate of the 3D position of the object center with respect to the gripper. The robot employs such information to move towards the object. The object is then grasped and lifted from the table (\textit{Grasp} phase). The microphone and the speech recognition toolkit are then employed again, to recognize an \enquote{open} verbal command to open the gripper and handover the object to the human (\textit{Handover} phase). After this, the robot moves back to a \textit{Pre-scan} position, and re-initiate the \textit{Scan} phase, ready to grasp other objects if asked by the user.

\section{Experiments} \label{sec:experiments}
We have validated the capabilities of the proposed gripper in a series of experiments to show both its high force and smart grasping capabilities.

\subsection{Grasping force Validation}\label{sec:experiments_mech}
While the standard specifications of most gripper devices usually include only the continuous/peak grasping force level and payload, in this work we perform an assessment study that considers a novel grasping payload evaluation procedure. In such tests, the robot arm dynamic motion and gripper thermal states are taken into account to reveal the true functional range of the gripper grasping force under static and dynamic motion of the grasped payload and under gripper actuation thermal stressing conditions.

\subsubsection{Grasping Force Capability}
\begin{figure}
    \centering
    \includegraphics[width=.8\linewidth, trim={0 1cm 7cm 2cm}, clip]{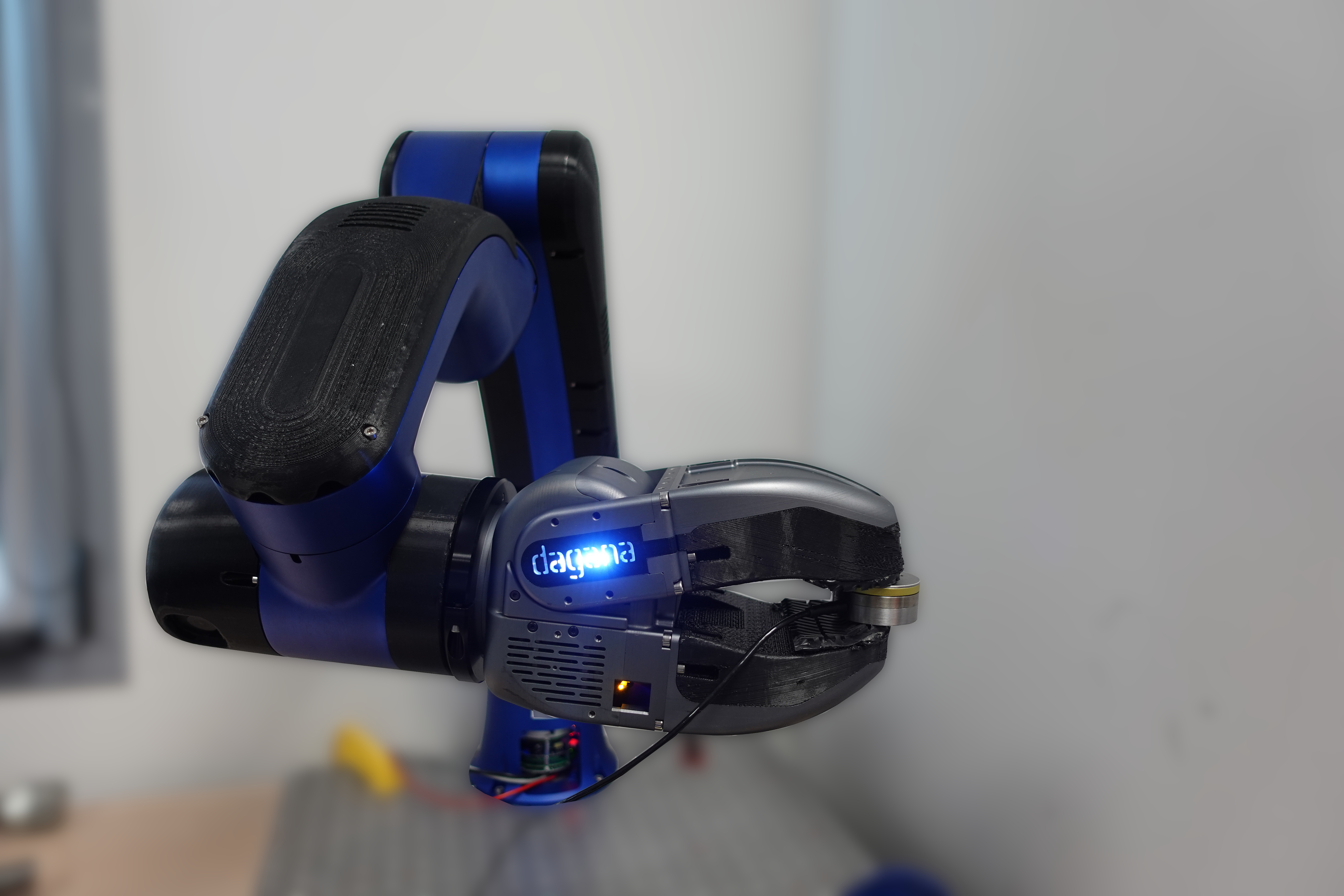}
    \caption{The gripper pressing the external force-torque sensor.}
    \label{fig:applied_force_img}
\end{figure}
\begin{figure}
    \centering
    \includegraphics[width=1\linewidth]{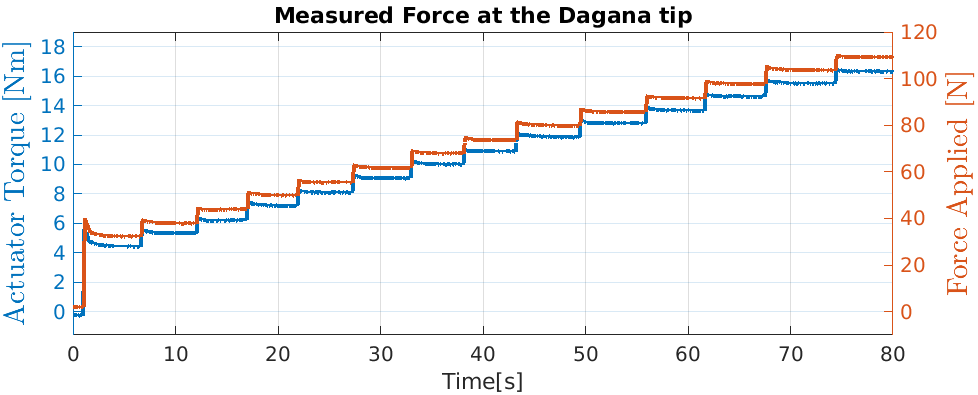}
    \caption{The grasping force generated by the gripper (right axis) as a consequence of its actuator torque increments (left axis). The actuation generates a peak grasping force of \SI{110}{\newton} at the tip of the jaws, which is in an approximate distance of \SI{0.15}{\meter} with respect to the center of the joint of moving jaw.}
    \label{fig:applied_force}
\end{figure}

This first experiment aims to assess the maximum grasping force that can be generated given the available torque of its actuation system. For this evaluation, we employed an external force-torque sensor (as shown in \figurename{}~\ref{fig:applied_force_img}) and commanded the gripper actuator to increase in small increments the generated torque up to its peak torque of (\SI{16.5}{\newton\meter}). The plot in \figurename{}~\ref{fig:applied_force} presents the resultant force measured by the force-torque sensor for different actuator torques showing a maximum grasping force capacity of \SI{110}{\newton} at the peak gripper actuation torque (\SI{16.5}{\newton\meter}).

\subsubsection{Thermal Stress Evaluation}
\begin{figure}
    \centering
    \includegraphics[width=\linewidth]{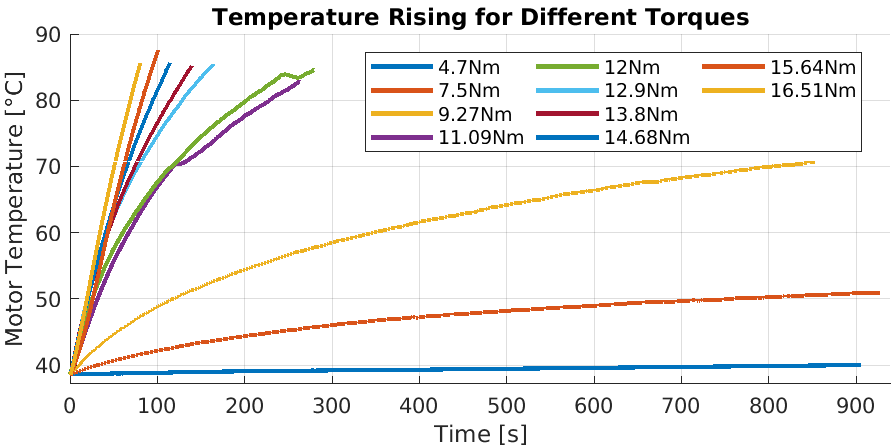}
    \caption{The curves show, for a given measured actuator torque, how the temperature of the actuator winding ($y$ axis) rises over time ($x$ axis), indicating how long the gripper can be used for delivering a desired torque.}
    \label{fig:temp}
\end{figure}

In a similar setup as in the previous test, we evaluated the thermal stress of the gripper actuator for different grasping force levels with the purpose to reveal the full picture of its grasping force range limits considering also the thermal constraint. While continuously commanding a specific static torque to the gripper actuator, the temperature of the motor winding was monitored. From \figurename{}~\ref{fig:temp}, it can be seen that the actuator of the gripper can generate a torque of approximately \SI{10}{\newton\meter} for a very long time, reaching a temperature of (\SI{70}{\degreeCelsius}) after approximately \SI{850}{\sec}. This is still largely below the temperature limit (\SI{130}{\degreeCelsius}) for the motor winding, according to the motor data sheet.  
Greater torque than \SI{10}{\newton\meter} are anyway achievable and tolerated for a satisfactory amount of time. In the experiment with higher torques, we stopped the thermal stress experiment when the actuator winding reaches a temperature of \SI{85}{\degreeCelsius}. It can be seen that, even for the peak torque of (\SI{16.5}{\newton\meter}) in which the gripper can generate a grasping force of \SI{110}{\newton}, the gripper can hold this peak grasping force for approximately \SI{80}{\sec}.

\subsubsection{Static Payload stress Evaluation}\label{sec:static_load}
\begin{figure}
    \centering
    \includegraphics[width=.9\linewidth, trim={10cm 2cm 0 0}, clip]{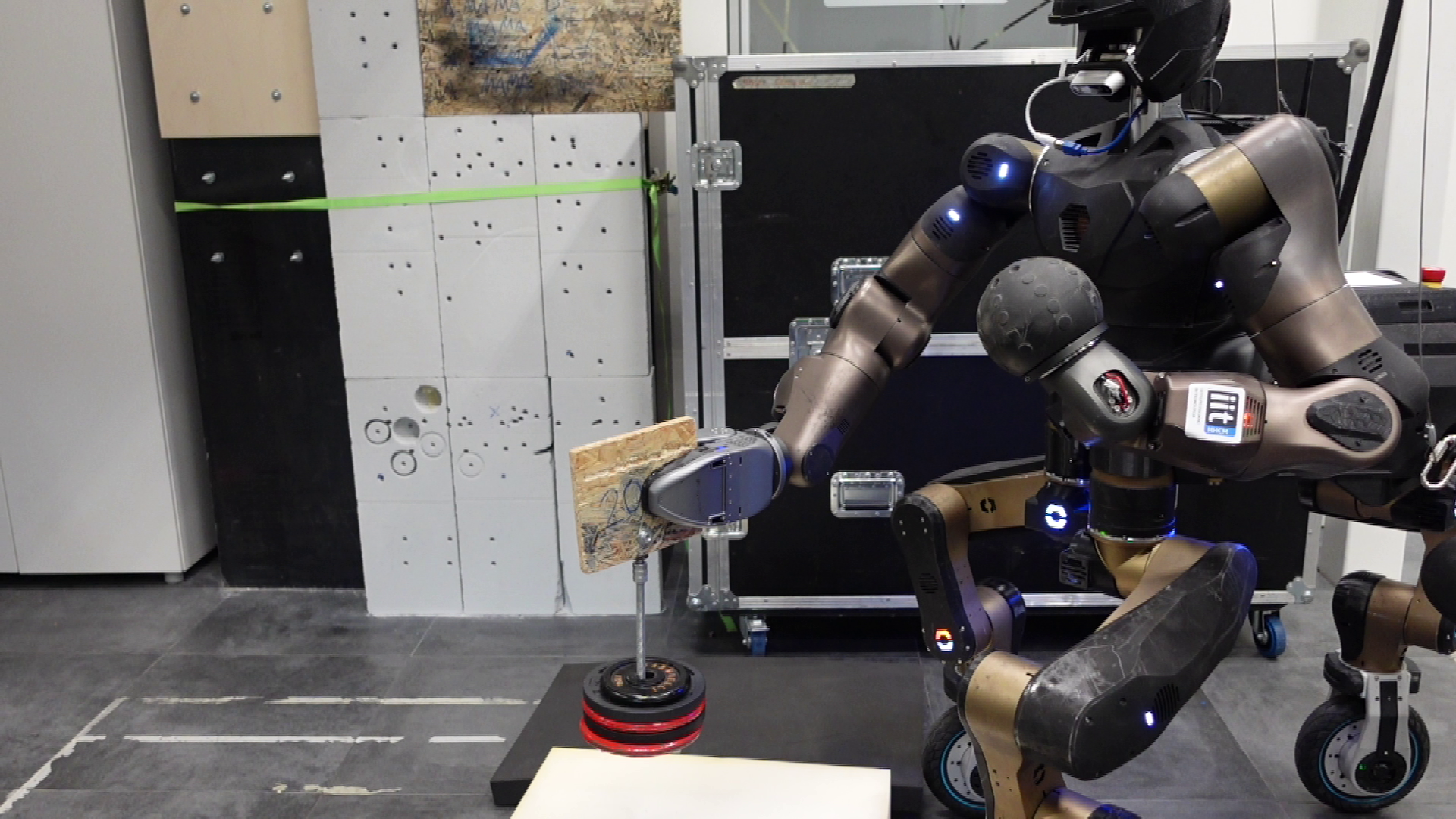}
    \caption{The CENTAURO robot holding a weight of \SI{9.390}{\kilogram} with the developed gripper, in one of the Static Payload stress tests.}
    \label{fig:static_load_centauro}
\end{figure}
\begin{figure}
    \centering
    \includegraphics[width=\linewidth]{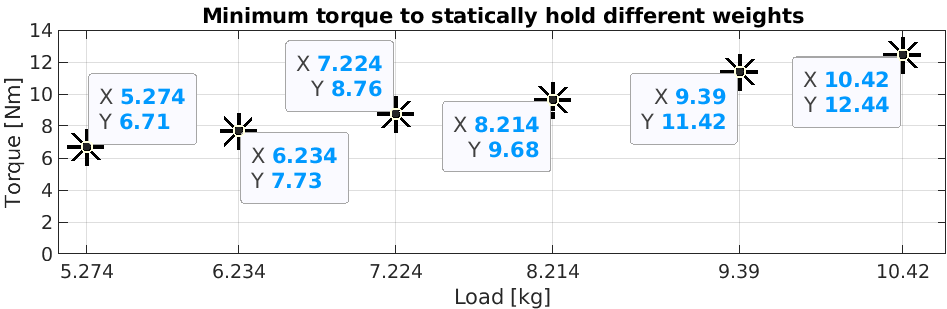}
    \caption{Plot from the Static Payload stress tests, showing the minimum torque ($y$ axis) necessary to hold a given weight ($x$ axis) without slippage.}
    \label{fig:static_load}
\end{figure}

This experiment aims to demonstrate the capability of the gripper in securely holding a payload statically, as shown in \figurename{}~\ref{fig:static_load_centauro}. Given a certain payload, we commanded a certain torque and slowly lowered it down until load's slippage was visually detected.
The object considered had a wooden handle and the jaws were equipped with rubber pads.
Comparing the temperature shown in \figurename{}~\ref{fig:temp}, we can notice that, for this pair of materials (wood-rubber), the gripper can statically hold a payload \SI{8}{\kilogram} for an indefinite amount of time, and up to \SI{10}{\kilogram} for approximately $3$ minutes.

\subsubsection{Dynamic Payload Stress Assessment}
\begin{figure}
    \centering
    \includegraphics[width=.48\linewidth, trim={18cm 4cm 0 0}, clip]{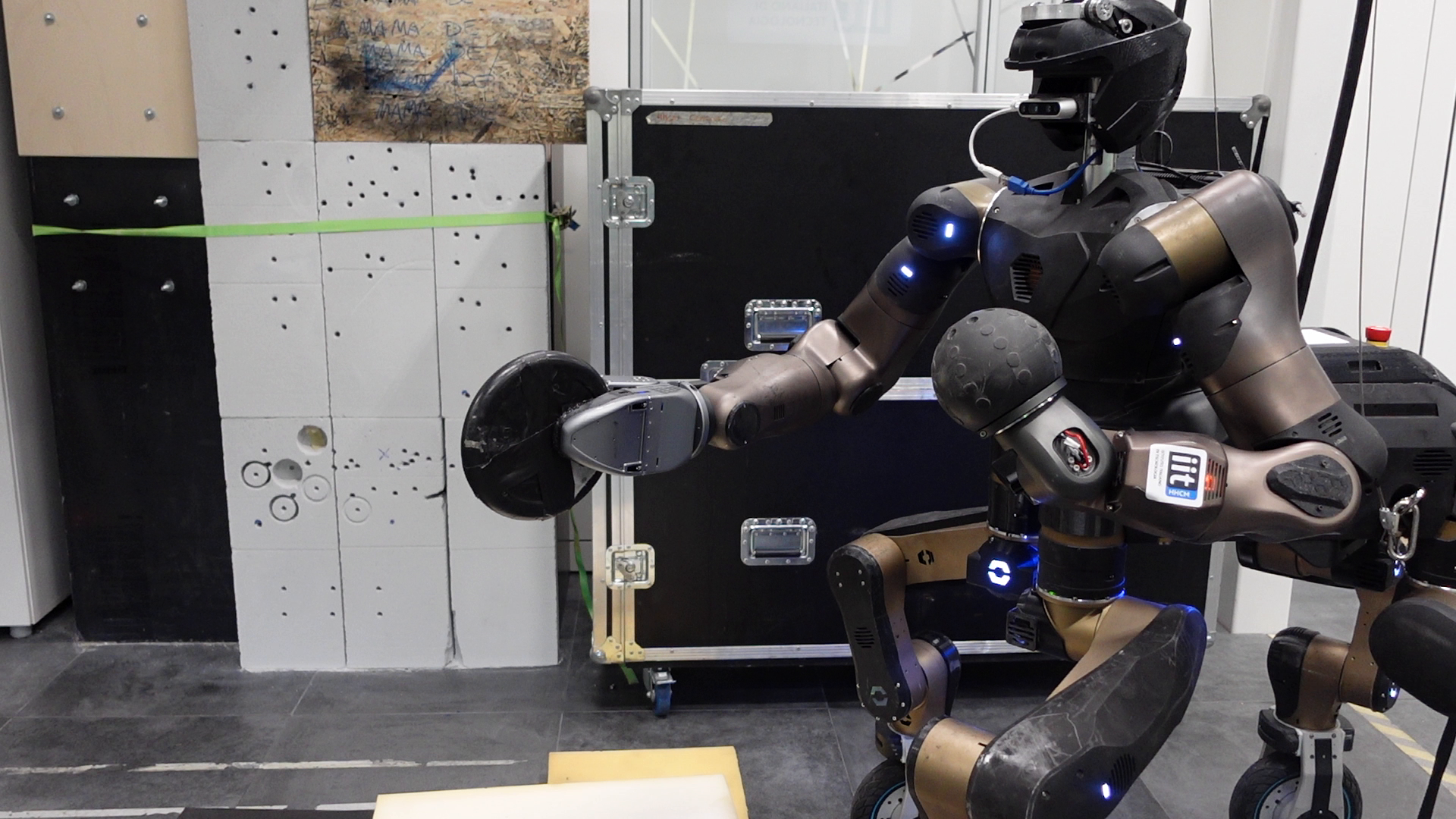}
    \includegraphics[width=.48\linewidth, trim={18cm 4cm 0 0}, clip]{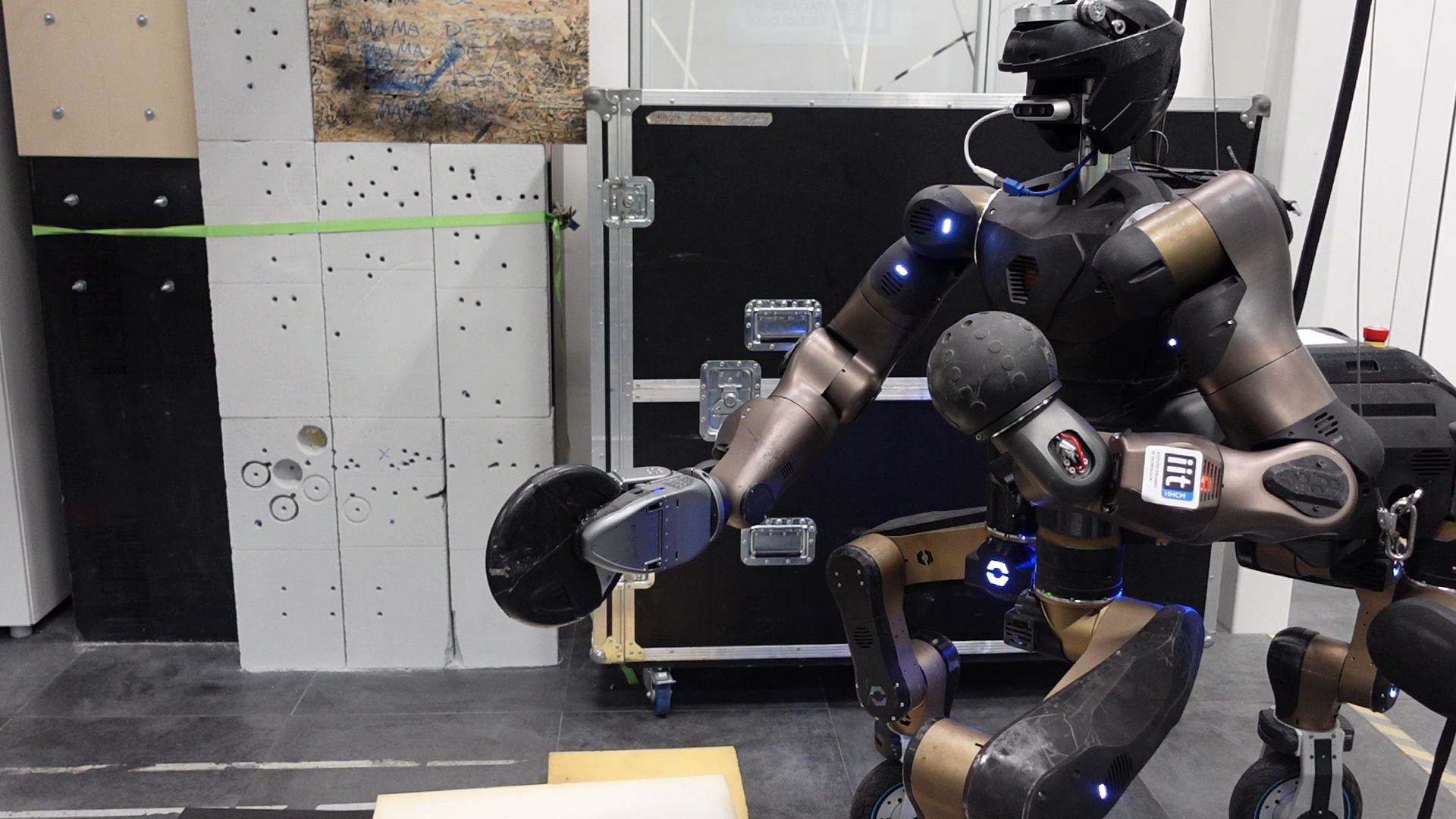}
    \caption{The CENTAURO robot moving the arm up and down while the gripper holds a payload of \SI{5.864}{\kilogram} in the Dynamic Payload stress tests.}
    \label{fig:dynamic_load_centauro}
\end{figure}
\begin{figure}
    \centering
    \includegraphics[width=\linewidth]{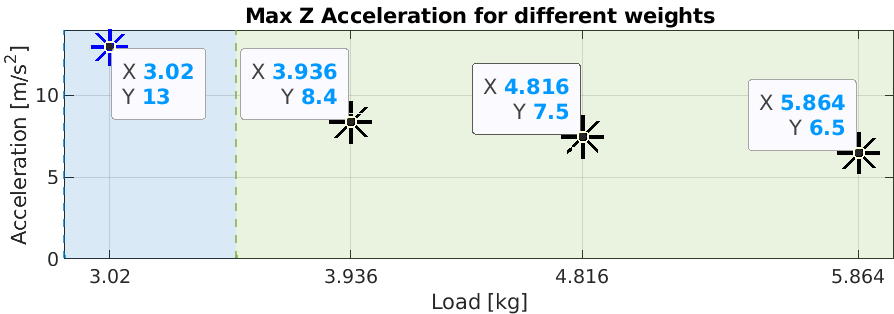}
    \caption{Results of the Dynamic Payload stress tests. For a given grasped payload ($x$ axis), it is shown the maximum acceleration achievable without triggering payload slipping. In the blue area, the acceleration level is limited by the CENTAURO arm joint limits, and not by the gripper's ability to hold the payload firmly at higher accelerations.}
    \vspace{-5px}
    \label{fig:dynamic_load}
\end{figure}

\begin{figure*}
    \centering
    \includegraphics[height=0.146\linewidth, trim={0 0 16.5cm 0},clip]{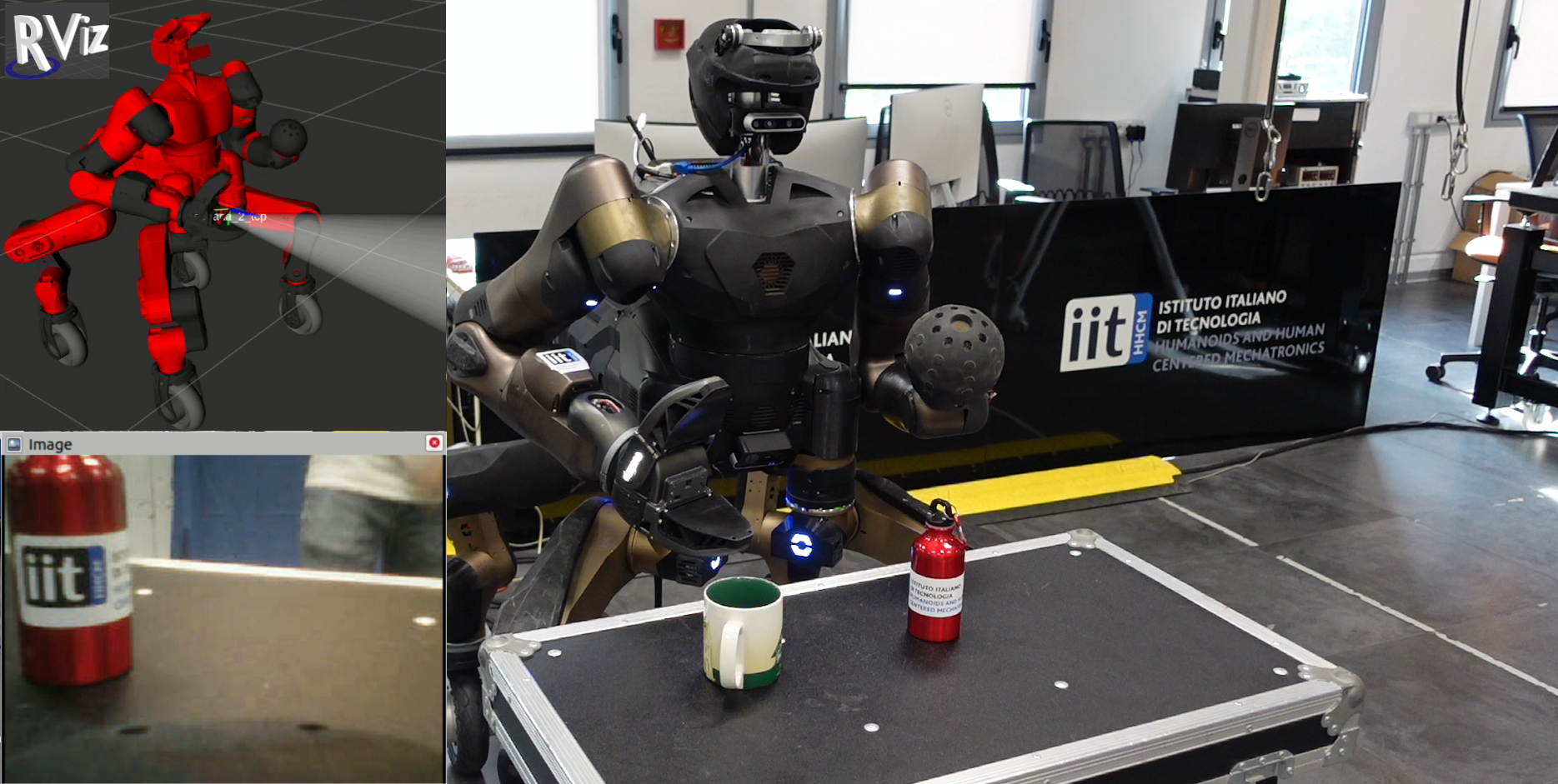}
    \includegraphics[height=0.146\linewidth, trim={0 0 16.5cm 0},clip]{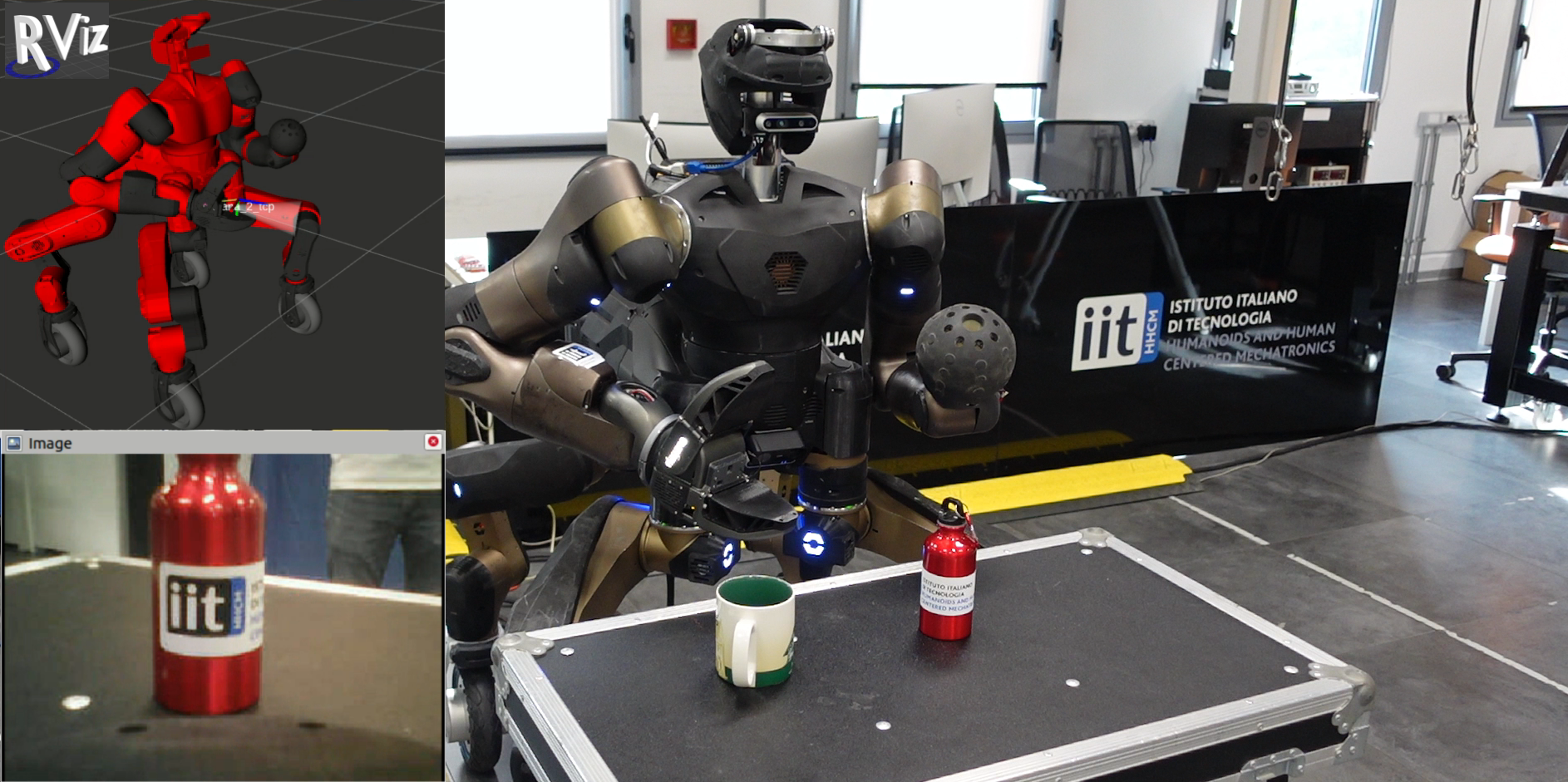}
    \includegraphics[height=0.146\linewidth, trim={0 0 16.5cm 0},clip]{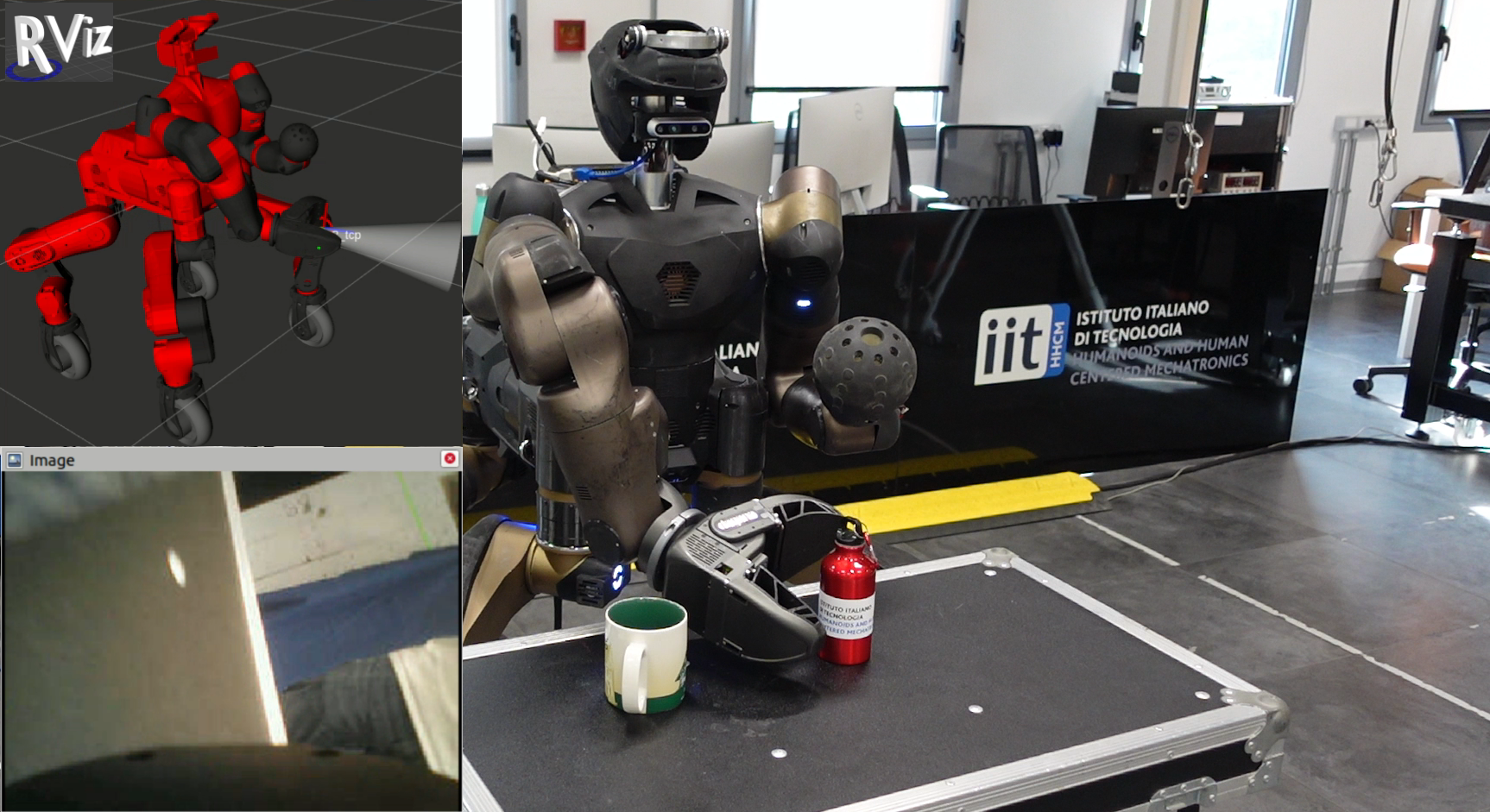}
    \includegraphics[height=0.146\linewidth, trim={0 0 9cm 0},clip]{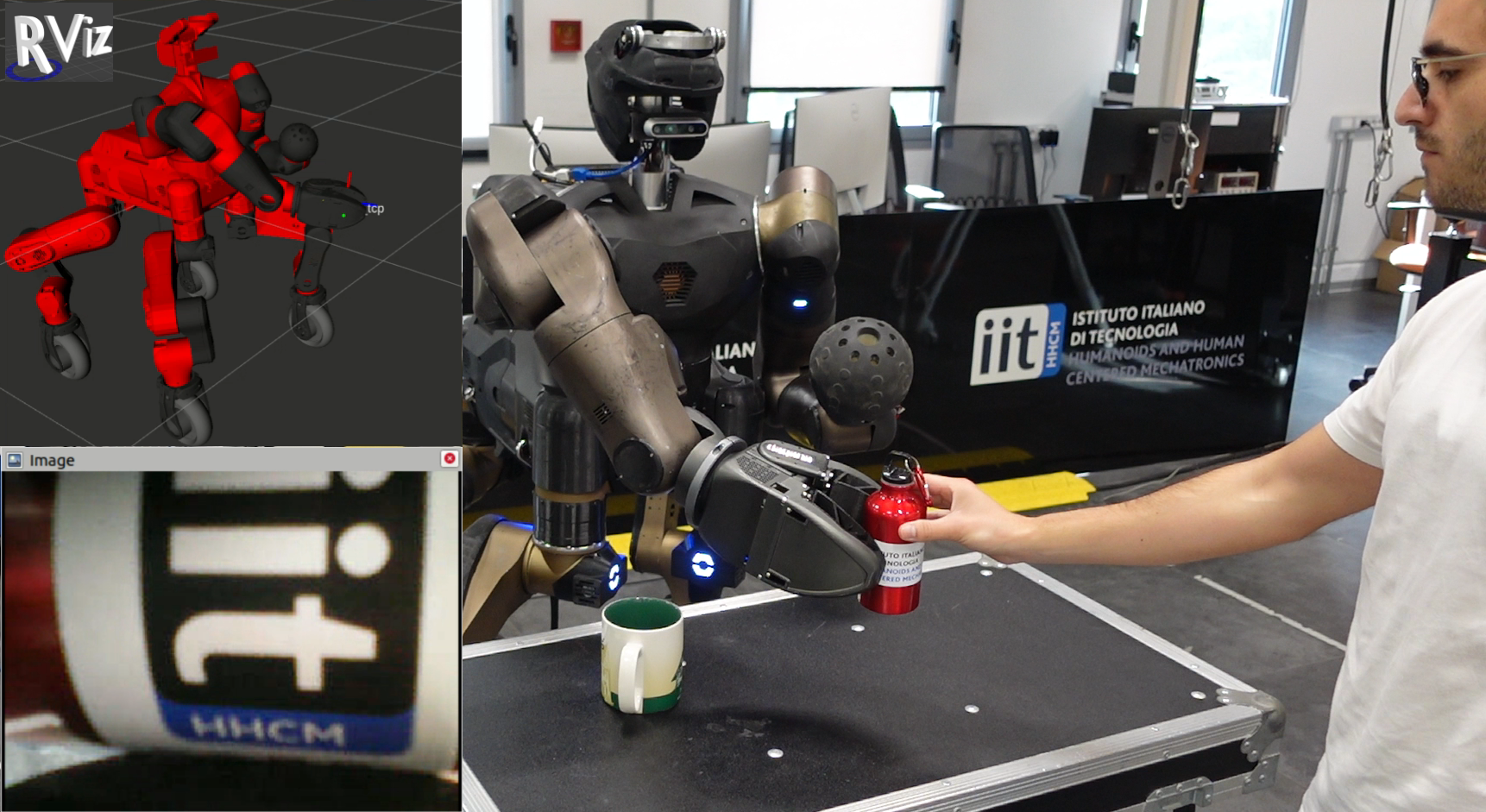}
    \includegraphics[height=0.146\linewidth, trim={0 0 16.5cm 0},clip]{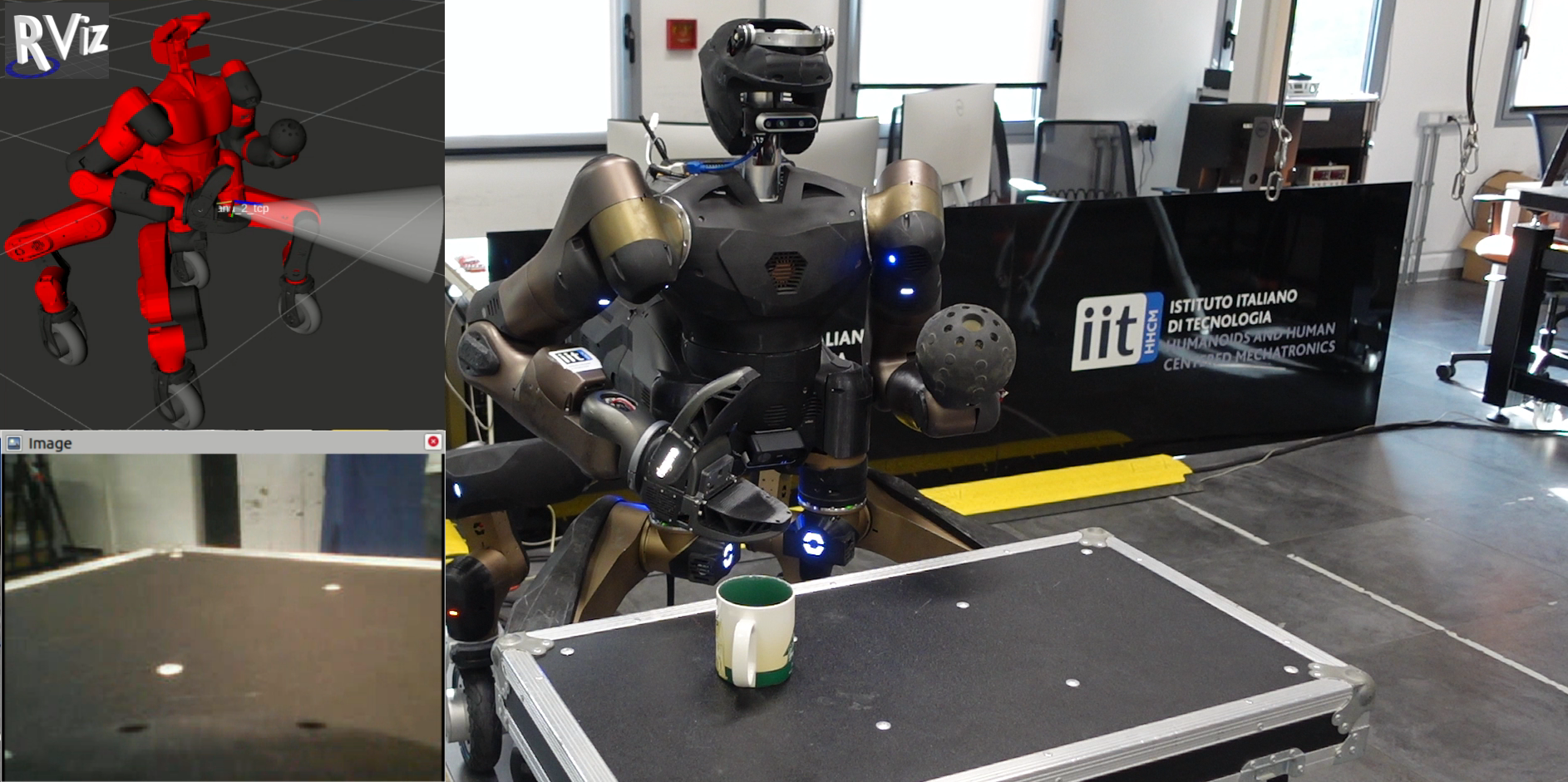}
    \caption{The CENTAURO robot in different phases of the Perception Driven Grasping experiment. From left to right: 1) The robot scans for objects with the embedded camera in the gripper; 2) The user commands to pick the "bottle", hence the robot aligns to it to grasp; 3) The robot grasps the object after measuring the distance with the range sensor; 4) The robot hands the object over to the user at the \enquote{open} command; 5) The robot returns in the scanning phase again.}
    \label{fig:grasping-exp}
\end{figure*}

\begin{figure}
    \centering
    \includegraphics[width=1\linewidth]{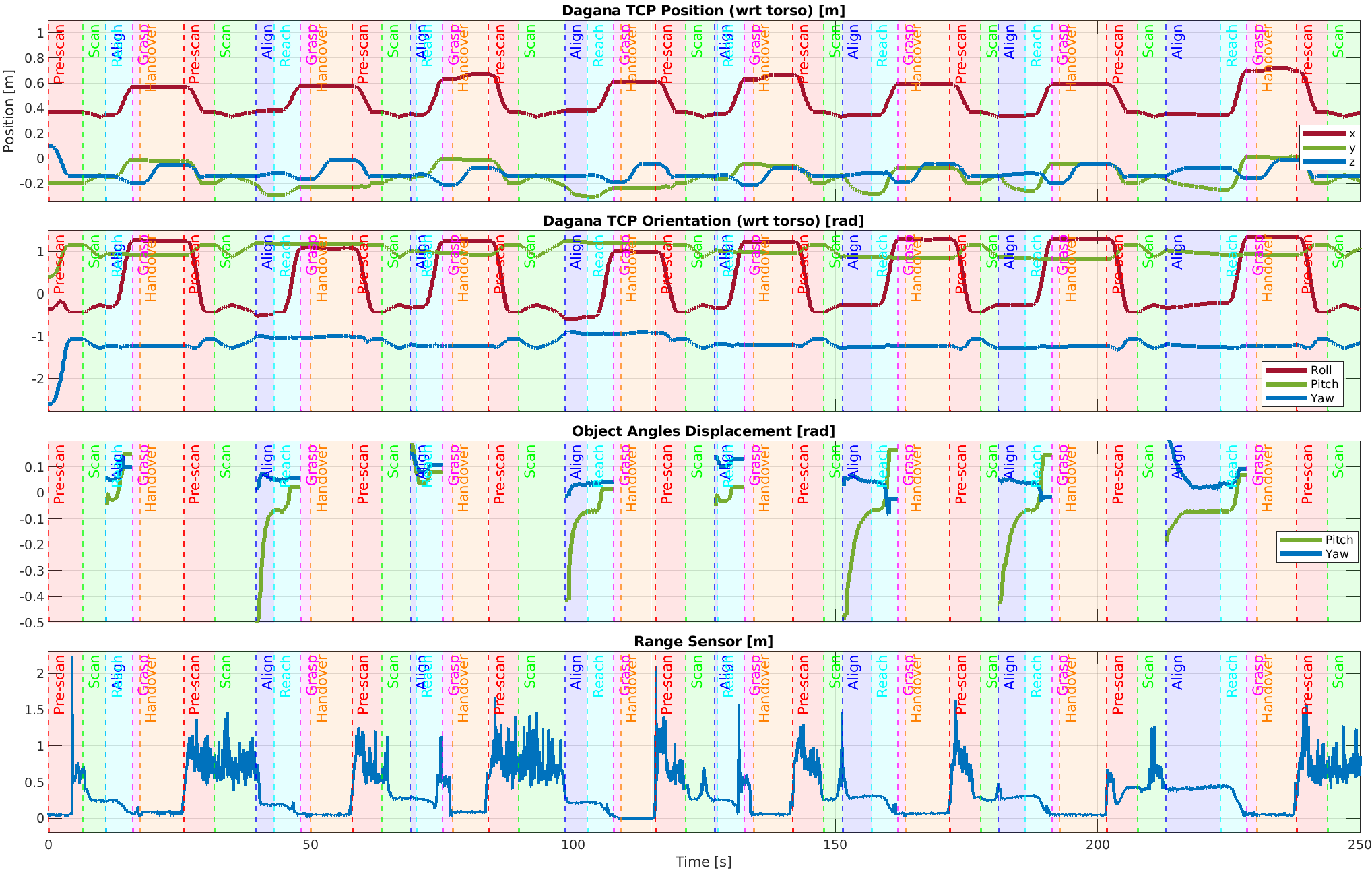}
    \caption{Data plots of the Perception Driven Grasping experiment. Coloured areas represent the states schematised in \figurename{}~\ref{fig:graspPerceptionPipeline}. The first two rows show the end-effector's TCP pose. The third plots the object displacements computed from the RGB images. The fourth depicts the ToF sensor data.}
    \label{fig:grasp-plot}
\end{figure}

We also conducted a dynamic grasped-payload experiment to further showcase the strength capacity of the gripper, as shown in \figurename{}~\ref{fig:dynamic_load_centauro}. In this experiment, the gripper holds a weight plate while the CENTAURO robot moves its arm up and down along the direction of gravity, with a certain Cartesian velocity. For a given weight, we increased the velocity of the motion in subsequent trials and collected the Cartesian acceleration, which, for the nature of the motion, generates peaks in the $z$ axis when the CENTAURO arm is at the top and the bottom point of the trajectory, that is when the change of motion direction suddenly occurs.
In these trials, the plates are directly held by the gripper. 
The results in \figurename{}~\ref{fig:dynamic_load} demonstrate, for a given payload, the maximum acceleration achievable, above which the payload slipping was observed. The torque of the gripper actuator was commanded at \SI{10.6}{\newton\meter}, considering this torque as the continuous ideal torque as understood in the previous thermal stress trials. It is important to note that for the \SI{3.02}{\kilogram} weight the acceleration limit was not imposed by the gripper's ability to firmly hold the payload, but by the limits of the CENTAURO arm joints, which prevented the generation of higher motion velocity and acceleration at the level of the end-effector. %
This experiment showed that the gripper can bear the range of the velocities/accelerations generated by the CENTAURO robot arm, ensuring stable grasping of considerable payload during dynamic arm motions.

\subsection{Perception Driven Grasping Validation}

In this experiment, we employ the perception-driven grasping functionalities of the gripper described in Section~\ref{sec:percGrasp}.
This validation aims to qualitatively demonstrate that the coupling between the RGB camera and the single-distance ToF sensor could be enough for picking some predetermined objects, e.g., a \enquote{bottle} or a \enquote{cup} as in our case. Furthermore, we show an example of the use of the embedded microphone to listen for vocal commands from the user directly from the gripper, without the need to have an additional microphone mounted on the robot body or worn by the user.

The experimental setup is made with the CENTAURO robot in front of a table with some objects to be grasped. In the images of \figurename{}\ref{fig:grasping-exp}, some phases of the perception-driven grasping pipeline are shown. After the user requests a particular object, the robot detects, reaches, and grasps it. At another user's request, the object is released and handed over to the user. The objects are picked up multiple times from the table from different positions, effectively showing  the use of the embedded multi-sensing module and the adaptability and robustness of the implemented pipeline that explores the multi-sensing module.

In the plots of \figurename{}~\ref{fig:grasp-plot}, data is shown for the entire experiment. The colored areas represent the subsequent different phases of the pipeline, from \textit{Pre-scan}, to \textit{Handover}, repeated multiple times. In the first two rows, the pose of the gripper's Tool Control Point (TCP) is shown. In the third row, the angles' displacement of the detected object are presented, reducing toward zero in the \textit{Align} phase. Finally, in the fourth row, the ToF sensor data is shown, employed in the \textit{Reach} phase to move the CENTAURO arm forward of an amount provided by the sensor.

The object orientation is assumed to be given for all the objects because we use objects geometrically distributed along the vertical direction, e.g., bottles on a table. These objects can be grasped by orienting the jaws of the gripper parallel to the objects' principal axis.
Although we did not delve into the grasping problem, this experiment proves qualitatively how the multi-sensing capabilities of the gripper can be used in conjunction with the grasping abilities to provide autonomous grasping skills and verbal user interactions without the necessity of any additional sensing device from the robot on which the gripper is mounted.

\section{Conclusions} \label{sec:conclusions}
This work introduced a modular gripper device that combines high grasping force capability with embedded multi-modal sensing features. After detailing the mechatronics of the gripper, we extensively validated its high grasping force capability identifying its operational limits under static and dynamic manipulation stress experiments that also monitored the thermal state of the gripper actuator. Furthermore, we made use of its multi-modal sensing equipment to perform a perception-driven grasping task. The presented gripper demonstrated a robust high-force manipulation system that is easy to customize according to the task at hand and allows the development of multi-modal perception-enhanced applications.  

In future works, more sophisticated perception-enhanced techniques will be developed. As an example, we will focus on pose estimation algorithms, using the distance sensor of the gripper for pose refinement in monocular 3D pose estimation~\cite{Ranftl2022} to solve the problem of monocular scale ambiguity. On what concerns the gripper mechatronics, they will be explored passive articulated, as well as soft, jaw modules that provide intrinsic adaptability and grasping force distribution to multiple contact points on the shape of the object.

\balance

\bibliographystyle{IEEEtranBST/IEEEtran}
\bibliography{IEEEtranBST/IEEEabrv,references.bib}

\begin{thebibliography}{10}
\providecommand{\url}[1]{#1}
\csname url@rmstyle\endcsname
\providecommand{\newblock}{\relax}
\providecommand{\bibinfo}[2]{#2}
\providecommand\BIBentrySTDinterwordspacing{\spaceskip=0pt\relax}
\providecommand\BIBentryALTinterwordstretchfactor{4}
\providecommand\BIBentryALTinterwordspacing{\spaceskip=\fontdimen2\font plus
\BIBentryALTinterwordstretchfactor\fontdimen3\font minus
  \fontdimen4\font\relax}
\providecommand\BIBforeignlanguage[2]{{%
\expandafter\ifx\csname l@#1\endcsname\relax
\typeout{** WARNING: IEEEtran.bst: No hyphenation pattern has been}%
\typeout{** loaded for the language `#1'. Using the pattern for}%
\typeout{** the default language instead.}%
\else
\language=\csname l@#1\endcsname
\fi
#2}}

\bibitem{brantner2021controlling}
G.~Brantner and O.~Khatib, ``{Controlling Ocean One: Human--robot collaboration
  for deep-sea manipulation},'' \emph{{J. of Field Robotics}}, vol.~38, no.~1,
  pp. 28--51, 2021.

\bibitem{brown2010universal}
E.~Brown \emph{et~al.}, ``{Universal Robotic Gripper based on the Jamming of
  Granular Material},'' \emph{{Proceedings of the National Academy of
  Sciences}}, vol. 107, no.~44, pp. 18\,809--18\,814, 2010.

\bibitem{langowski2020soft}
J.~Langowski, P.~Sharma, and A.~L. Shoushtari, ``{In the soft grip of
  nature},'' \emph{{Science Robotics}}, vol.~5, no.~49, p. eabd9120, 2020.

\bibitem{kashiri2019centauro}
N.~Kashiri \emph{et~al.}, ``{CENTAURO: A Hybrid Locomotion and High Power
  Resilient Manipulation Platform},'' \emph{{IEEE} Robot. Autom. Lett.},
  vol.~4, no.~2, pp. 1595--1602, 2019.

\bibitem{gripperDesignSurvey}
J.~Hernandez \emph{et~al.}, ``Current designs of robotic arm grippers: A
  comprehensive systematic review,'' \emph{Robotics}, vol.~12, no.~1, 2023.

\bibitem{samadikhoshkho2019brief}
Z.~Samadikhoshkho, K.~Zareinia, and F.~Janabi-Sharifi, ``{A brief review on
  robotic grippers classifications},'' in \emph{{IEEE Canadian Conf. of
  Electrical and Computer Engineering}}, 2019, pp. 1--4.

\bibitem{Negrello2020}
F.~Negrello, H.~S. Stuart, and M.~G. Catalano, ``{Hands in the Real World},''
  \emph{Frontiers in Robotics and AI}, vol.~6, 2020.

\bibitem{torielli2023ros}
D.~Torielli, L.~Bertoni, F.~Fusaro, N.~Tsagarakis, and L.~Muratore, ``{ROS}
  end-effector: A hardware-agnostic software and control framework for robotic
  end-effectors,'' \emph{J. of Intelligent \& Robotic Systems}, vol. 108,
  no.~4, p.~70, 2023.

\bibitem{ma2016m}
R.~R. Ma, A.~Spiers, and A.~M. Dollar, ``{M 2 gripper: Extending the Dexterity
  of a Simple, Underactuated Gripper},'' in \emph{{Adv.in Reconfigurable
  Mechanisms and Robots II}}, 2016, pp. 795--805.

\bibitem{Piazza2019}
C.~Piazza, G.~Grioli, M.~Catalano, and A.~Bicchi, ``{A Century of Robotic
  Hands},'' \emph{Annual Review of Control, Robotics, and Autonomous Systems},
  vol.~2, no.~1, pp. 1--32, 2019.

\bibitem{Catalano2016}
M.~Catalano \emph{et~al.}, \emph{From Soft to Adaptive Synergies: The Pisa/IIT
  SoftHand}, 2016, ch.~8, pp. 101--125.

\bibitem{RBOHand22016}
R.~Deimel and O.~Brock, ``A novel type of compliant and underactuated robotic
  hand for dexterous grasping,'' \emph{The Int. J. of Robotics Research},
  vol.~35, no. 1-3, pp. 161--185, 2016.

\bibitem{ren2018heri}
Z.~Ren, N.~Kashiri, C.~Zhou, and N.~G. Tsagarakis, ``{HERI II: A Robust and
  Flexible Robotic Hand based on Modular Finger design and Under Actuation
  Principles},'' in \emph{{IEEE/RSJ} Int. Conf. Intell. Robots Syst.}, 2018,
  pp. 1449--1455.

\bibitem{Liu2020}
C.-H. Liu, F.-M. Chung, Y.~Chen, C.-H. Chiu, and T.-L. Chen, ``Optimal design
  of a motor-driven three-finger soft robotic gripper,'' \emph{IEEE/ASME Trans.
  Mechatronics}, vol.~25, no.~4, pp. 1830--1840, 2020.

\bibitem{AllegroHand2012}
J.-H. Bae \emph{et~al.}, ``Development of a low cost anthropomorphic robot hand
  with high capability,'' in \emph{{IEEE/RSJ} Int. Conf. Intell. Robots Syst.},
  2012, pp. 4776--4782.

\bibitem{Ruehl2014}
S.~W. Ruehl \emph{et~al.}, ``{Experimental evaluation of the schunk 5-Finger
  gripping hand for grasping tasks},'' \emph{IEEE Int. Conf. on Robotics and
  Biomimetics}, pp. 2465--2470, 2014.

\bibitem{shadowhand2014}
G.~Walck, U.~Cupcic, T.~Duran, and V.~Perdereau, ``A case study of ros software
  re-usability for dexterous in-hand manipulation,'' \emph{J. of Software
  Engineering for Robotics}, 05 2014.

\bibitem{wu2023back}
X.~Wu, H.~Hua, C.~Zhao, N.~Shi, and Z.~Wu, ``A back-drivable rotational force
  actuator for adaptive grasping,'' in \emph{Actuators}, vol.~12, no.~7, 2023,
  p. 267.

\bibitem{Chao2023}
C.~Liu, A.~Moncada, H.~Matusik, D.~I. Erus, and D.~Rus, ``A modular
  bio-inspired robotic hand with high sensitivity,'' in \emph{{IEEE} Int. Conf.
  Soft Robotics}, 2023, pp. 1--7.

\bibitem{maeda2022f1}
G.~Maeda, N.~Fukaya, and S.-i. Maeda, ``{F1 hand: A versatile fixed-finger
  gripper for delicate teleoperation and autonomous grasping},'' \emph{{IEEE}
  Robot. Autom. Lett.}, vol.~7, no.~3, pp. 6734--6741, 2022.

\bibitem{telegenov2015low}
K.~Telegenov, Y.~Tlegenov, and A.~Shintemirov, ``{A Low-Cost Open-Source
  3-D-Printed Three-Finger Gripper Platform for Research and Educational
  Purposes},'' \emph{IEEE access}, vol.~3, pp. 638--647, 2015.

\bibitem{bostondynamics2024}
{Boston Dynamics}, ``{Spot Arm},''
  https://bostondynamics.com/wp-content/uploads/2020/10/spot-arm.pdf, accessed:
  15/06/2024.

\bibitem{KAPPASSOV2015195}
Z.~Kappassov, J.-A. Corrales, and V.~Perdereau, ``Tactile sensing in dexterous
  robot hands — review,'' \emph{Robotics and Autonomous Systems}, vol.~74,
  pp. 195--220, 2015.

\bibitem{Romano2011}
J.~M. Romano, K.~Hsiao, G.~Niemeyer, S.~Chitta, and K.~J. Kuchenbecker,
  ``Human-inspired robotic grasp control with tactile sensing,'' \emph{IEEE
  Trans. on Robotics}, vol.~27, no.~6, pp. 1067--1079, 2011.

\bibitem{yamaguchi2016combining}
A.~Yamaguchi and C.~G. Atkeson, ``{Combining finger vision and optical tactile
  sensing: Reducing and handling errors while cutting vegetables},'' in
  \emph{{IEEE-RAS Int. Conf. Humanoid Robots}}, 2016, pp. 1045--1051.

\bibitem{Narita2020}
T.~Narita \emph{et~al.}, ``Theoretical derivation and realization of adaptive
  grasping based on rotational incipient slip detection,'' in \emph{{IEEE} Int.
  Conf. Robot. Autom.}, 2020, pp. 531--537.

\bibitem{cigliano2015robotic}
P.~Cigliano, V.~Lippiello, F.~Ruggiero, and B.~Siciliano, ``{Robotic ball
  catching with an eye-in-hand single-camera system},'' \emph{{IEEE Trans. on
  Control Systems Technology}}, vol.~23, no.~5, pp. 1657--1671, 2015.

\bibitem{Hundhausen2021}
F.~Hundhausen, R.~Grimm, L.~Stieber, and T.~Asfour, ``Fast reactive grasping
  with in-finger vision and in-hand fpga-accelerated cnns,'' in
  \emph{{IEEE/RSJ} Int. Conf. Intell. Robots Syst.}, 2021, pp. 6825--6832.

\bibitem{Cirillo2021}
A.~Cirillo, G.~Laudante, and S.~Pirozzi, ``Proximity sensor for thin wire
  recognition and manipulation,'' \emph{Machines}, vol.~9, no.~9, 2021.

\bibitem{Mejia2024}
J.~Mejia, V.~Dean, T.~Hellebrekers, and A.~Gupta, ``Hearing touch: Audio-visual
  pretraining for contact-rich manipulation,'' in \emph{IEEE Int. Conf. on
  Robotics and Automation}, 2024.

\bibitem{Denoun2023}
B.~Denoun, M.~Hansard, B.~León, and L.~Jamone, ``Statistical stratification
  and benchmarking of robotic grasping performance,'' \emph{IEEE Trans. on
  Robotics}, vol.~39, no.~6, pp. 4539--4551, 2023.

\bibitem{Sotiropoulos2018}
P.~Sotiropoulos \emph{et~al.}, ``A benchmarking framework for systematic
  evaluation of compliant under-actuated soft end effectors in an industrial
  context,'' in \emph{IEEE-RAS Int. Conf. Humanoid Robots}, 2018, pp. 280--283.

\bibitem{Angus2023}
A.~B. Clark, L.~Cramphorn-Neal, M.~Rachowiecki, and A.~Gregg-Smith, ``Household
  clothing set and benchmarks for characterising end-effector cloth
  manipulation,'' in \emph{{IEEE} Int. Conf. Robot. Autom.}, 2023, pp.
  9211--9217.

\bibitem{Falco2015}
J.~Falco, K.~Van~Wyk, S.~Liu, and S.~Carpin, ``Grasping the performance:
  Facilitating replicable performance measures via benchmarking and
  standardized methodologies,'' \emph{{IEEE} Robot. Autom. Mag.}, vol.~22,
  no.~4, pp. 125--136, 2015.

\bibitem{XBot2}
A.~Laurenzi, D.~Antonucci, N.~Tsagarakis, and L.~Muratore, ``The xbot2
  real-time middleware for robotics,'' \emph{Robot. Auton. Syst.}, vol. 163, p.
  104379, 2023.

\bibitem{ROS}
\BIBentryALTinterwordspacing
M.~Quigley \emph{et~al.}, ``{ROS Robot Operating System [Computer software]}.''
  [Online]. Available: \url{https://www.ros.org/}
\BIBentrySTDinterwordspacing

\bibitem{vosk}
\BIBentryALTinterwordspacing
Alphacephei, ``{VOSK Offline Speech Recognition API}.'' [Online]. Available:
  \url{https://alphacephei.com/vosk/}
\BIBentrySTDinterwordspacing

\bibitem{lin2014microsoft}
T.-Y. Lin \emph{et~al.}, ``Microsoft {COCO}: Common objects in context,'' in
  \emph{Computer Vision -- ECCV 2014}, 2014, pp. 740--755.

\bibitem{redmon2016you}
J.~Redmon, S.~Divvala, R.~Girshick, and A.~Farhadi, ``You only look once:
  Unified, real-time object detection,'' in \emph{Conf. on Computer Vision and
  Pattern Recognition}, 2016.

\bibitem{Ranftl2022}
R.~Ranftl, K.~Lasinger, D.~Hafner, K.~Schindler, and V.~Koltun, ``Towards
  robust monocular depth estimation: Mixing datasets for zero-shot
  cross-dataset transfer,'' \emph{IEEE Trans. on Pattern Analysis and Machine
  Intelligence}, 2022.

\end{thebibliography}

\end{document}